\documentclass[runningheads]{llncs}
\usepackage{graphicx}
\usepackage{comment}

\usepackage{booktabs}       
\usepackage{amsfonts}       
\usepackage{nicefrac}       
\usepackage{microtype}  

\usepackage{graphicx}
\usepackage{multirow}
\usepackage{array}
\usepackage{rotating} 
\usepackage{amssymb}
\usepackage{amsmath}
\usepackage{wrapfig}
\usepackage{tabularx}
\usepackage{pifont}

\usepackage{amsmath,amssymb} 
\usepackage{color}

\begin{document}
\pagestyle{headings}
\mainmatter
\def\ECCVSubNumber{5818}  

\title{Label-Driven Reconstruction for Domain Adaptation in Semantic Segmentation} 

\titlerunning{Label-Driven Reconstruction for Domain Adaptation}
%
\author{Jinyu Yang\orcidID{0000-0002-7004-3570} \and
Weizhi An \and
Sheng Wang \and
Xinliang Zhu \and 
Chaochao Yan\orcidID{0000-0003-1237-8978} \and 
Junzhou Huang\orcidID{0000-0002-9548-1227}}
\authorrunning{Yang J., An W., Wang S., Zhu X., Yan C., Huang J.}
%
\institute{University of Texas at Arlington, Texas, USA \\
\email{\{jinyu.yang, weizhi.an, sheng.wang, xinliang.zhu, chaochao.yan\}@mavs.uta.edu}, jzhuang@uta.edu}
\maketitle

\begin{abstract}
Unsupervised domain adaptation enables to alleviate the need for pixel-wise annotation in the semantic segmentation. One of the most common strategies is to translate images from the source domain to the target domain and then align their marginal distributions in the feature space using adversarial learning. However, source-to-target translation enlarges the bias in translated images and introduces extra computations, owing to the dominant data size of the source domain. Furthermore, consistency of the joint distribution in source and target domains cannot be guaranteed through global feature alignment. Here, we present an innovative framework, designed to mitigate the image translation bias and align cross-domain features with the same category. This is achieved by 1) performing the target-to-source translation and 2) reconstructing both source and target images from their predicted labels. Extensive experiments on adapting from synthetic to real urban scene understanding demonstrate that our framework competes favorably against existing state-of-the-art methods.

\end{abstract}	
	
	\section{Introduction}
	Deep Convolutional Neural Networks (DCNNs) have demonstrated impressive achievements in computer vision tasks, such as image recognition \cite{he2016deep}, object detection \cite{girshick2015fast}, and semantic segmentation \cite{long2015fully}. As one of the most fundamental tasks, semantic segmentation predicts pixel-level semantic labels for given images. It plays an extremely important role in autonomous agent applications such as self-driving techniques.

	Existing supervised semantic segmentation methods, however, largely rely on pixel-wise annotations which require tremendous time and labor efforts. To overcome this limitation, publicly available synthetic datasets (e.g., GTA \cite{richter2016playing} and SYNTHIA \cite{ros2016synthia}) which are densely-annotated, have been considered recently. Nevertheless, the most obvious drawback of such a strategy is the poor knowledge generalization caused by domain shift issues (e.g., appearance and spatial layout differences), giving rise to dramatic performance degradation when directly applying models learned from synthetic data to real-world data of interest. In consequence, domain adaptation has been exploited in recent studies for cross-domain semantic segmentation, where the most common strategy is to learn domain-invariant representations by minimizing distribution discrepancy between source and target domains \cite{zhang2017curriculum,lian2019constructing}, designing a new loss function \cite{zhu2018penalizing}, considering depth information \cite{chen2019learning,vu2019dada}, or alternatively generating highly confident pseudo labels and re-training models with these labels through a self-training manner \cite{zou2018unsupervised,li2019bidirectional,huang2020contextual,pan2020unsupervised,wang2020differential,kim2020learning,yang2020fda}. Following the advances of Generative Adversarial Nets (GAN) \cite{goodfellow2014generative}, adversarial learning has been used to match cross-domain representations by minimizing an adversarial loss on the source and target representations \cite{hoffman2016fcns,luo2019significance,luo2019taking,Du_2019_ICCV}, or adapting structured output space across two domains \cite{tsai2018learning,li2019bidirectional}. Recent studies further consider the pixel-level (e.g., texture and lighting) domain shift to enforce source and target images to be domain-invariant in terms of visual appearance \cite{zhang2018fully,hoffman2017cycada,wu2018dcan,murez2018image,chen2019crdoco,yang2020context}. This is achieved by translating images from the source domain to the target domain by using image-to-image translation models such as CycleGAN \cite{zhu2017unpaired} and UNIT \cite{liu2017unsupervised}.

	Despite these painstaking efforts, we are still far from being able to fully adapt cross-domain knowledge mainly stemming from two limitations. First, adversarial-based image-to-image translation introduces inevitable bias to the translated images, as we cannot fully guarantee that the translated source domain $ \mathcal{F}(\mathcal{X}_s) $ is identical to the target domain $ \mathcal{X}_t $ ($ \mathcal{X}_s $ and $ \mathcal{X}_t $ denote two domains, and $ \mathcal{F}$ indicates an image-to-image translation model). This limitation is especially harmful to the source-to-target translation \cite{zhang2018fully,hoffman2017cycada,wu2018dcan,murez2018image,li2019bidirectional}, since the data size of the source domain is much larger than the target domain in most of domain adaptation problems. Moreover, source-to-target translation is more computationally expensive than target-to-source translation. Second, simply aligning cross-domain representations in the feature space \cite{hoffman2016fcns,hoffman2017cycada,tsai2018learning} ignores the joint distribution shift (i.e., $ \mathcal{P}(G(\mathcal{X}_s), Y_s)  \neq \mathcal{P}(G(\mathcal{X}_t), Y_t) $, where $ G $ is used for feature extraction, while $ Y_s $ and $ Y_t $ indicate ground truth labels). These limitations give rise to severe false positive and false negative issues in the target prediction. This problem can get even worse when there is a significant discrepancy in layout or structure between the source and target domains, such as adapting from synthetic to real urban traffic scenes.
	
	In this paper, we propose an innovative domain adaptation framework for semantic segmentation. The key idea is to reduce the image translation bias and align cross-domain feature representations through image reconstruction. As opposed to performing source-to-target translation \cite{hoffman2017cycada,wu2018dcan,li2019bidirectional}, for the first time, we conduct the target-to-source translation to make target images indistinguishable from source images. This enables us to substantially reduce the bias in translated images and allows us to use original source images and their corresponding ground truth to train a segmentation network. Compared to the source-to-target translation, our method is also much more efficient. Besides, a reconstruction network is designed to reconstruct both source and target images from their predicted labels. It is noteworthy that we reconstruct images directly from the label space, rather than the feature space as reported in previous studies. This is essential to guide the segmentation network by penalizing the reconstructed image that semantically deviates from the corresponding input image. Most importantly, this strategy enforces cross-domain features with the same category close to each other.
	
	\begin{figure*}[t]
		\begin{center}
			\includegraphics[width=1.0\linewidth]{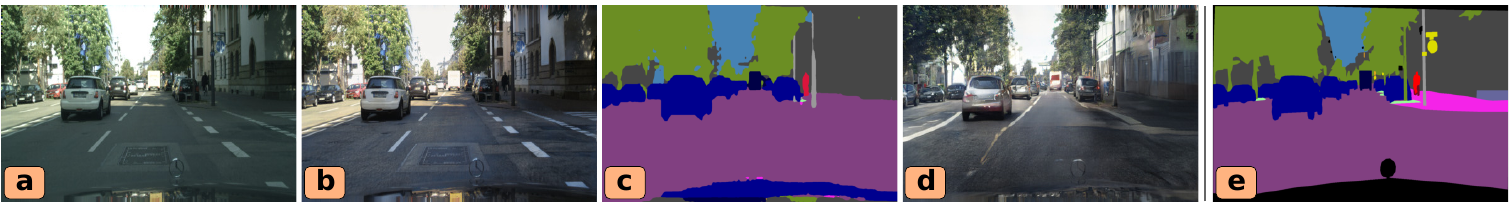}
		\end{center}
		\caption{An example of our method on synthetic-to-real urban scene adaptation. Given a target-domain (or real) image (a), we first make target-to-source translation to obtain source-like (or synthetic) image (b), and then perform segmentation on these translated images. Our method improves the segmentation accuracy in the target domain by reconstructing both source and target images from their predicted labels (c). (d) illustrates the image reconstructed from (c), while (e) indicates the ground truth label.}
		\label{fig:target_semantic}
	\end{figure*}
	
	
	The performance of our method is evaluated on synthetic-to-real scenarios of urban scene understanding, i.e., GTA5 to Cityscapes and SYNTHIA to Cityscapes. Our results demonstrate that the proposed method achieves significant improvements compared with existing methods. Figure~\ref{fig:target_semantic} demonstrates an example of our model in adapting cross-domain knowledge in semantic segmentation tasks and reconstructing the input image from its output label. We also carry out comprehensive ablation studies to analyze the effectiveness of each component in our framework.
	
	The contribution of this paper is threefold. 
	\begin{itemize}
	    \item For the first time, we propose and investigate the target-to-source translation in domain adaptation. It reduces the image translation bias and is more computationally efficient compared to the widely-used source-to-target translation.
	    \item To enforce semantic consistency, we introduce a label-driven reconstruction module that reconstructs both source and target images from their predicted labels.
	    \item Extensive experiments show that our method achieves the new state-of-the-art performance on adapting synthetic-to-real semantic segmentation.
	\end{itemize}

	
	\section{Related Work}
	\paragraph{\textbf{Semantic Segmentation}} Recent achievements in semantic segmentation mainly benefit from the technical advances of DCNNs, especially the emergence of Fully Convolutional Network (FCN) \cite{long2015fully}. By adapting and extending contemporary deep classification architectures fully convolutionally, FCN enables pixel-wise semantic prediction for any arbitrary-sized inputs and has been widely recognized as one of the benchmark methods in this field. Numerous methods inspired by FCN were then proposed to further enhance segmentation accuracy, which have exhibited distinct performance improvement on the well-known datasets (e.g., PASCAL VOC 2012 \cite{pascal-voc-2012} and Cityscapes \cite{cordts2016cityscapes}) \cite{chen2014semantic,liu2015semantic,zhao2017pyramid,chen2018deeplab,chen2018searching}.
	
	However, such methods heavily rely on human-annotated, pixel-level segmentation masks, which require extremely expensive labeling efforts \cite{cordts2016cityscapes}. In consequence, weakly-supervised methods, which are based on easily obtained annotations (e.g., bounding boxes and image-level tags), were proposed to alleviate the need for effort-consuming labeling \cite{dai2015boxsup,pinheiro2015image}. Another alternative is to resort to freely-available synthetic datasets (e.g., GTA5 \cite{richter2016playing} and SYNTHIA \cite{ros2016synthia}) with pixel-level semantic annotations. However, models learned on synthetic datasets suffer from significant performance degradation when directly applied to the real datasets of interest, mainly owing to the domain shift issue. 
	
	\paragraph{{\bf Domain Adaptation}} Domain adaptation aims to mitigate the domain discrepancy between a source and a target domain, which can be further divided into supervised adaptation, semi-supervised adaptation, and unsupervised adaptation, depending on the availability of labels in the target domain. The term unsupervised domain adaptation refers to the scenario where target labels are unavailable and have been extensively studied \cite{long2015learning,tzeng2014deep,ganin2014unsupervised,tzeng2015simultaneous,tzeng2017adversarial,ying2018transfer,zhang2019collaborative}.
	
	
	
	Recent publications have highlighted the complementary role of pixel-level and representation-level adaptation in semantic segmentation \cite{hoffman2017cycada,murez2017image,zhang2018fully,wu2018dcan,chen2019learning}, where the pixel-level adaptation is mainly achieved by translating images from the source domain to the target domain (source-to-target translation). Specifically, unpaired image-to-image translation is used in CyCADA \cite{hoffman2017cycada} to achieve pixel-level adaptation by restricting cycle-consistency. Similarly, FCAN achieves the image translation by combining the image content in the source domain and the "style" from the target domain \cite{zhang2018fully}. I2IAdapt \cite{murez2017image} further considers to align source and target representations based on an image-to-image translation strategy, attempting to adapt domain shift. Instead of using the adversarial learning for image translation, DCAN performs source-to-target translation by leveraging target images for channel-wise alignment \cite{wu2018dcan}. Driven by the fact that geometry and semantics are coordinated with each other, GIO-Ada augments the standard image translation network by integrating geometric information \cite{chen2019learning}. However, source-to-target translation introduces substantial bias to the translated images, given that the size of the source domain is usually much larger than the target domain. To address this problem, we propose the first-of-its-kind target-to-source image translation to reduce pixel-level domain discrepancy. Compared to the source-to-target translation, it is more computationally efficient and enables us to remove the uncertainty by training the segmentation network with original source images and their corresponding labels.
	
	
	Motivated by the observation that cross-domain images (e.g., GTA5 and Cityscapes) often share tremendous structural similarities, ASN \cite{tsai2018learning} adapts structured output based on the adversarial learning. The strength of this method is its ability to provide weak supervision to target images by enforcing target outs to be indistinguishable from source outputs. However, it is limited to the scenario where two domains have a huge layout discrepancy, resulting in meaningless predictions for target images. To address this limitation, we further enforce the semantic consistency between target images and their predicted labels through a reconstruction network.
	
	\setlength\intextsep{0pt}
    \begin{wrapfigure}{r}{0.55\textwidth}
		\begin{center}
			\includegraphics[width=1\linewidth]{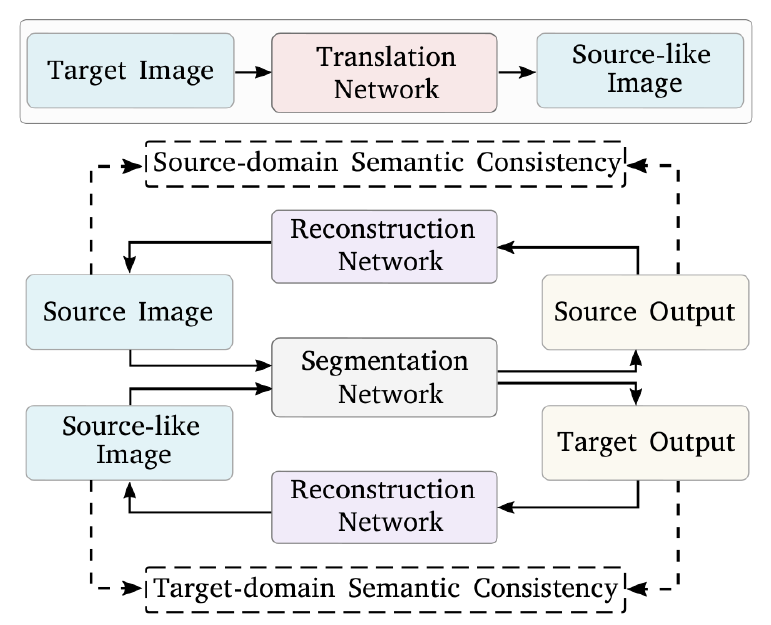}
		\end{center}
		\caption{An overview of our framework.}
		\label{fig:overview}
	\end{wrapfigure}
	
	Inspired by the self-training, \cite{li2019bidirectional,huang2020contextual,pan2020unsupervised,wang2020differential,kim2020learning,yang2020fda} generate pseudo labels for target images and then re-training the segmentation model with these labels. It outperforms the existing methods by a large margin. However, such a strategy underestimates the side effect of pseudo labels that are incorrectly predicted. As a consequence, the segmentation model fails to increasingly improve itself using these wrong ground truth. Instead, our method reconstructs source and target input images from the label space to ensure these outputs are semantically correct. The image-to-image translation network in \cite{li2019bidirectional} uses a reconstruction loss and a perceptual loss to maintain the semantic consistency between the input image and the reconstruction from the translated image. Different from \cite{li2019bidirectional}, we design a cycle-reconstruction loss in our reconstruction network to enforce the semantic consistency between the input image and the reconstruction from the predicted label.
	
	Reconstruction-based strategy for unsupervised domain adaptation has received considerable attention recently \cite{ghifary2016deep,bousmalis2016domain}. The key idea is to reconstruct input images from their feature representations to ensure that the segmentation model can learn useful information. Chang et al. \cite{chang2019all} follow a similar idea to first disentangle images into the domain-invariant structure and domain-specific texture representations, and then reconstruct input images. LSD-seg \cite{sankaranarayanan2018learning} first reconstructs images from the feature space, and then apply a discriminator to the reconstructed images. Rather than performing reconstruction from feature representations, we reconstruct both source and target images from their predicted labels.
	
	
	
	\section{Algorithm}
	\subsection{Overview}
	
	The overall design of our framework is illustrated in Figure~\ref{fig:overview}, mainly containing three complementary modules: a translation network $ \mathcal{F} $, a segmentation network $ G $, and a reconstruction network $ \mathcal{M} $. Given a set of source domain images $ \mathcal{X}_s $ with labels $ Y_s $ and a set of target domain images $ \mathcal{X}_t $ without any annotations. Our goal is to train $ G $ to predict accurate pixel-level labels for $ \mathcal{X}_t $. To achieve this, we first use $ \mathcal{F} $ to adapt pixel-level knowledge between $ \mathcal{X}_t $ and $ \mathcal{X}_s $ by translating $ \mathcal{X}_t $ to source-like images $ \mathcal{X}_{t{\rightarrow}s} $. This is different from existing prevalent methods that translate images from the source domain to the target domain. $ \mathcal{X}_s $ and $ \mathcal{X}_{t{\rightarrow}s} $ are then fed into $ G $ to predict their segmentation outputs $ G(\mathcal{X}_s) $ and $ G(\mathcal{X}_{t{\rightarrow}s}) $, respectively. To further enforce semantic consistency of both source and target domains, $ \mathcal{M} $ is then applied to reconstruct $ \mathcal{X}_s $ and $ \mathcal{X}_{t{\rightarrow}s} $ from their predicted labels. Specifically, a cycle-reconstruction loss is proposed to measure the reconstruction error, which enforces the semantic consistency and further guides segmentation network to predict more accurate segmentation outputs. 
	
	
	\subsection{Target-to-source Translation}
	
\begin{figure*}[t]
		\begin{center}
			\includegraphics[width=1.0\linewidth]{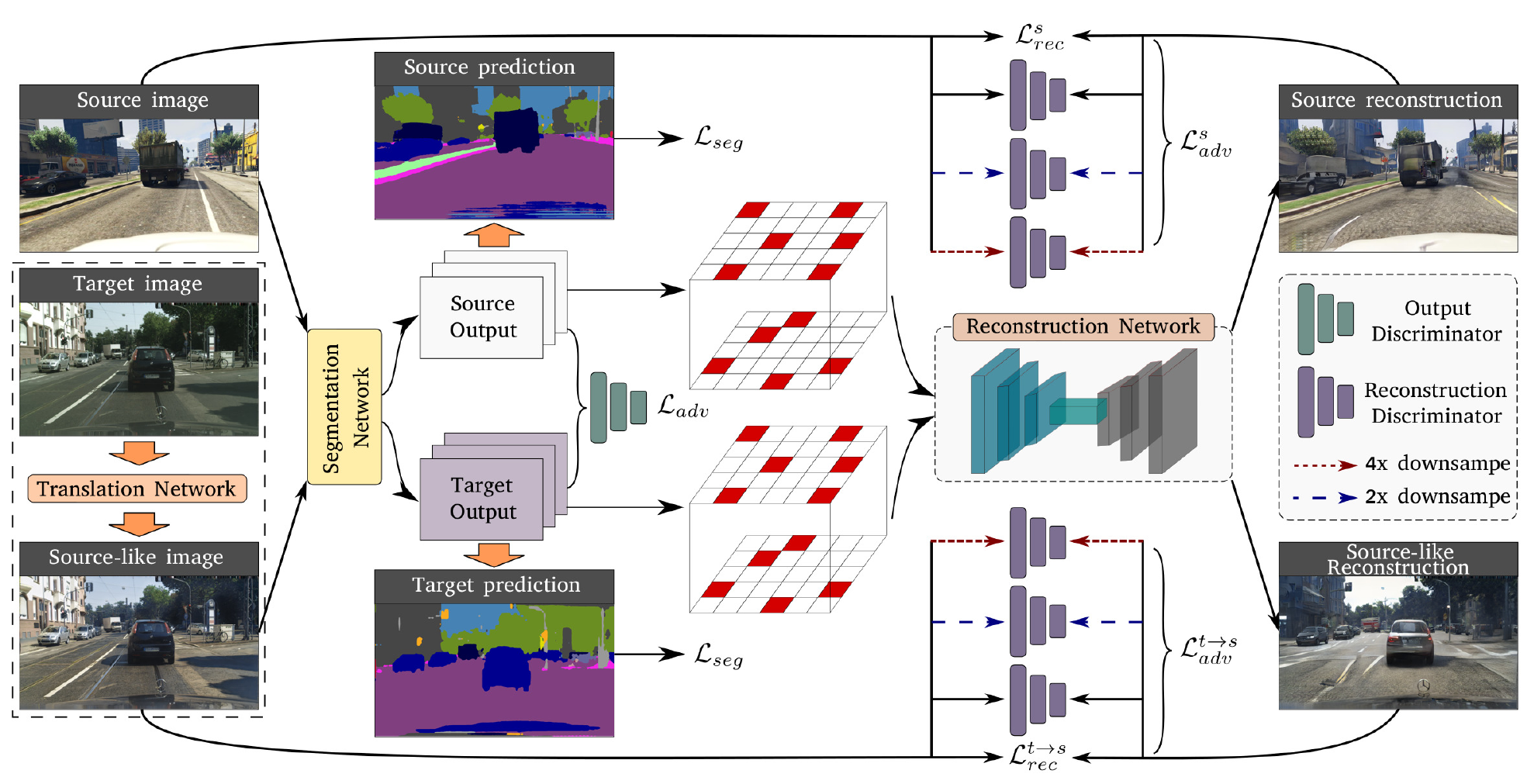}
		\end{center}
		\caption{Schematic overview of our framework which has three modules: (i) a translation network for pixel-level discrepancy reduction by translating target images to source-like images, where source-like images are indistinguishable from source images, (ii) a segmentation network that predicts segmentation outputs for source images and source-like images, and (iii) a reconstruction network for reconstructing source and source-like images from their corresponding label space.}
		\label{fig:framework}
	\end{figure*}

	We first perform the image-to-image translation to reduce the pixel-level discrepancy between source and target domains. As opposed to the source-to-target translation reported in previous domain adaptation methods, we conduct the target-to-source translation through an unsupervised image translation network (Figure~\ref{fig:framework}). Our goal is to learn a mapping $ \mathcal{F} $ : $ \mathcal{X}_t{\rightarrow}\mathcal{X}_s $ such that the distribution of images from $ \mathcal{F}(\mathcal{X}_t) $ is indistinguishable from the distribution of $ \mathcal{X}_s $. As a counterpart, the inverse mapping $ \mathcal{F}^{-1} $ : $ \mathcal{X}_s{\rightarrow}\mathcal{X}_t $, which maps images from $ \mathcal{X}_s $ to $ \mathcal{X}_t $, is introduced to prevent the mode collapse issue \cite{goodfellow2016nips}. Two adversarial discriminators $ \mathcal{D}_t $ and $ \mathcal{D}_s $ are employed for distribution match, where $ \mathcal{D}_t $ enforces indistinguishable distribution between $ \mathcal{F}(\mathcal{X}_t) $ and $ \mathcal{X}_s $, and $ \mathcal{D}_s $ encourages indistinguishable distribution between $ \mathcal{F}^{-1}(\mathcal{X}_s) $ and $ \mathcal{X}_t $ (More details can be found in the \textbf{Supplementary}).

	Based on the trained model $ \mathcal{F} $, we first translate images from $ \mathcal{X}_t $ to source-like images $ \mathcal{X}_{t{\rightarrow}s} = \mathcal{F}(\mathcal{X}_t) $. Specifically, each image in $ \mathcal{X}_{t{\rightarrow}s} $ preserves the same content as the corresponding image in $ \mathcal{X}_t $ while demonstrating the common style (e.g., texture and lighting) as $ \mathcal{X}_s $. $ \mathcal{X}_s $ and $ \mathcal{X}_{t{\rightarrow}s} $ are then fed into a segmentation network for semantic label prediction.
	
	Compared to translating images from the source domain to the target domain, the target-to-source translation has three benefits. First, it allows full supervision on the source domain by training the segmentation network with original source images and their corresponding labels. Second, it enables to reduce the bias in translated images. Third, it is computationally efficient lying in the fact that $ |\mathcal{X}_t| \ll |\mathcal{X}_s| $.

	
	
	
	
	\subsection{Semantic Segmentation}
	
	Given that source-like images $ \mathcal{X}_{t{\rightarrow}s} $ preserves all semantic information from $ \mathcal{X}_t $, we apply a shared segmentation network $ G $ to $ \mathcal{X}_s $ and $ \mathcal{X}_{t{\rightarrow}s} $ to predict their segmentation outputs with the loss function given by,
	\begin{equation} \label{eq:1}
	\begin{aligned}
	\mathcal{L}_{G} = {} &
	\mathcal{L}_{seg}(G(\mathcal{X}_s), Y_s) + \mathcal{L}_{seg}(G(\mathcal{X}_{t{\rightarrow}s}), Y_t^{ssl}) + \\ &
	\lambda \mathcal{L}_{adv}(G(\mathcal{X}_s), G(\mathcal{X}_{t{\rightarrow}s})),
	\end{aligned}
	\end{equation}
	where $ \mathcal{L}_{seg} $ indicates the typical segmentation objective, $ Y_t^{ssl} $ is pseudo labels of $ \mathcal{X}_{t{\rightarrow}s} $ which is derived from \cite{li2019bidirectional}, $ \mathcal{L}_{adv}(G(\mathcal{X}_s), G(\mathcal{X}_{t{\rightarrow}s})) $ is an adversarial loss, and $ \lambda $ leverages the importance of these losses. Specifically, $ \mathcal{L}_{adv}(G(\mathcal{X}_s), G(\mathcal{X}_{t{\rightarrow}s})) $ is defined as,
	\begin{equation}
	\begin{aligned}
	\mathcal{L}_{adv}(G(\mathcal{X}_s), G(\mathcal{X}_{t{\rightarrow}s})) = {} & 
	\mathop{\mathbb{E}}[log{D(G(\mathcal{X}_s))}] + \\ &
	\mathop{\mathbb{E}}[log{(1 - D(G(\mathcal{X}_{t{\rightarrow}s})))}],
	\end{aligned}
	\end{equation}
	which enforces $ G $ to learn domain-invariant features by confusing the discriminator $ D $. It is noteworthy that we regard the segmentation outputs $ G(\mathcal{X}_s) $ and $ G(\mathcal{X}_{t{\rightarrow}s}) $ as features in our study. This is based on the observation that $ \mathcal{X}_s $ and $ \mathcal{X}_{t{\rightarrow}s} $ share significant similarities in terms of spatial layouts and structures \cite{tsai2018learning}. 
	

	\begin{figure}[t]
		\begin{center}
			\includegraphics[width=0.85\linewidth]{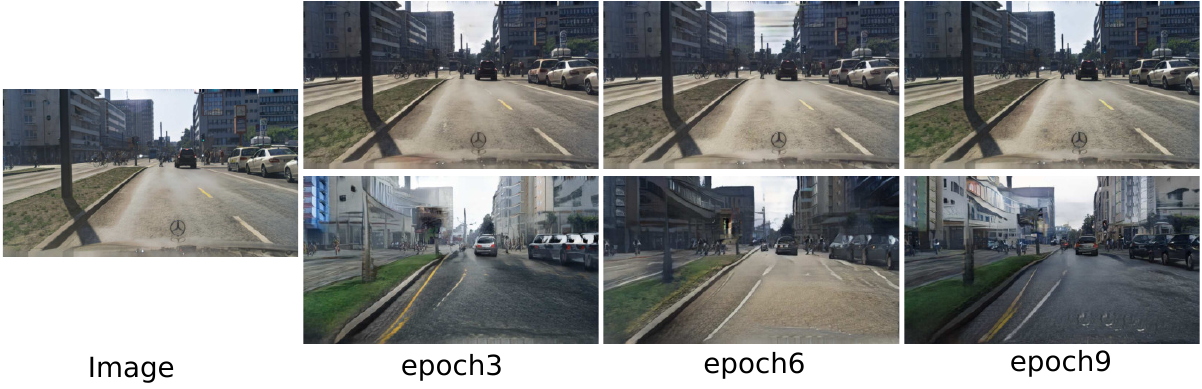}
		\end{center}
		\caption{A comparison between the image reconstruction from feature space and label space (ours). For each input image (first column), the first and second row indicate the images reconstructed from the feature and label space, respectively.}
		\label{fig:reconstruction}
	\end{figure}
	
	\subsection{Image Reconstruction from the Label Space}
	
	To encourage $ G $ to generate segmentation outputs that are semantic consistent, we introduce a reconstruction network $ \mathcal{M} $ to reconstruct $ \mathcal{X}_{\phi} $ from $ G(\mathcal{X}_{\phi}) \in \mathbb{R}^{H_{\phi} \times W_{\phi} \times C} $, where ($ H_{\phi}, W_{\phi} $) indicates image size, $ C $ represents the number of label classes, and the subscript $ \phi $ can be either $ s $ or $ t{\rightarrow}s $ to denote the source or the target domain. However, directly reconstructing images from the feature space fails to provide semantic consistency constraint to $ G $. On the one hand, $ G(\mathcal{X}_{\phi}) $ encodes rich information which makes the image reconstruction quite straightforward. As illustrated in Figure~\ref{fig:reconstruction}, in just a few epochs, the reconstructed images derived from $ \mathcal{M} $ are almost identical to the input images. On the other hand, to enforce cross-domain features with the same category close to each other, it is essential to perform the reconstruction based on the label space. Unfortunately, $ G(\mathcal{X}_{\phi}) $ lies in the feature space instead. To overcome these limitations, the most clear-cut way is to convert $ G(\mathcal{X}_{\phi}) $ to have zeros everywhere except where the index of each maximum value in the last dimension. Doing so formulates the categorical representation of the predicted label that corresponds to $ G(\mathcal{X}_{\phi}) $. Nevertheless, such conversion is non-differentiable and cannot be trained using standard backpropagation.

	Driven by the softmax action selection which is commonly used in the reinforcement learning, we apply Boltzmann distributed probabilities to approximate the semantic label map of $ G(\mathcal{X}_{\phi}) $, which is defined as,
	\begin{equation}
	{\Omega_{\phi}}^{(h, w, i)} = \frac{exp({G(\mathcal{X}_{\phi})}^{(h, w, i)} / \tau)}{\sum_{j=1}^{c} exp({G(\mathcal{X}_{\phi})}^{(h, w, j)} / \tau)},
	\end{equation}
	where $ \tau $ is a temperature parameter. This conversion is continuous and differentiable, therefore, we use $ \mathcal{M} $ to reconstruct input images $ \mathcal{X}_{\phi} $ from $ \Omega_{\phi} $ (Figure~\ref{fig:reconstruction}).
	
	
	
	
	
	
	
	
	
	To synthesize high-resolution images from the semantic label map, we use conditional GANs \cite{isola2017image} to model the conditional distribution of $ \mathcal{X}_{\phi} $ given $ \Omega_{\phi} $. To this end, we introduce $ \mathcal{M} $ and multi-scale domain discriminators $ D_{k} $ for $ k = 1,2,3 $. $ \mathcal{M} $ is designed to reconstruct $ \mathcal{X}_{\phi} $ from $ \Omega_{\phi} $, and $ D_{k} $ aims to distinguish $ \mathcal{X}_{\phi} $ from $ \mathcal{M}(\Omega_{\phi}) $. Specifically, $ \mathcal{M} $ follows the architecture proposed in \cite{johnson2016perceptual}, while $ D_{k} $ is based on PatchGAN \cite{isola2017image} that penalizes structure at the scale of image patches. All $ D_{k} $ follow the same network architecture. Besides $ \mathcal{X}_{\phi} $ and $ \mathcal{M}(\Omega_{\phi}) $ themselves, they are downsampled by a factor of 2 and 4 to obtain pyramid of 3 scales for $ D_{1} $, $ D_{2} $, and $ D_{3} $, respectively. It is worth mentioning that $ D_{k} $ is essential to differentiate real and reconstructed images with high resolution \cite{wang2017high}, owing to its ability in providing large receptive field. The objective function is given by,
	\begin{equation}
	\begin{aligned}
	{\mathcal{L}_{adv}^{\phi}} = {} &
	\sum\nolimits_{k=1}^{3}[\mathop{\mathbb{E}}[log{D_k(\Omega_{\phi}, \mathcal{X}_{\phi})}] + \\ & \mathop{\mathbb{E}}[log{(1 - D_k(\Omega_{\phi}, \mathcal{M}(\Omega_{\phi})))}]]
	\end{aligned}
	\end{equation}
	
	To further enforce semantic consistency between $ \mathcal{X}_{\phi} $ and $ \mathcal{M}(\Omega_{\phi}) $, we introduce a cycle-reconstruction loss $ \mathcal{L}_{rec}^{\phi} $ to match their feature representations. $ \mathcal{L}_{rec}^{\phi} $ contains a VGG perceptual loss and a discriminator feature matching loss, which is defined as,
	\begin{equation} \label{eq:recons}
	\begin{aligned}
	\mathcal{L}_{rec}^{\phi} = {} &
	\mathop{\mathbb{E}}\sum_{m=1}^{M}[||V^{(m)}(\mathcal{M}(\Omega_{\phi})) - V^{(m)}(\mathcal{X}_{\phi})||_1] + \\ &
	\mathop{\mathbb{E}}\sum_{k=1}^{3}\sum_{n=1}^{N}[||D^{(n)}_k(\Omega_{\phi}, \mathcal{X}_{\phi})) - D^{(n)}_k(\Omega_{\phi}, \mathcal{M}(\Omega_{\phi}))||_1]
	\end{aligned}
	\end{equation} 
	where $ V $ is a VGG19-based model for extracting high-level perceptual information \cite{johnson2016perceptual}, $ M $ and $ N $ represent the total number of layers in $ V $ and $ D_{k} $ for matching intermediate representations. Note that $ \mathcal{L}_{rec}^{\phi} $ penalizes $ \Omega_{\phi} $ when it deviates from the corresponding image $ \mathcal{X}_{\phi} $ in terms of semantic consistency. In this way, $ \mathcal{M} $ enables to map features from $ \mathcal{X}_{t{\rightarrow}s} $ closer to the features from $ \mathcal{X}_s $ with the same label. 
	
	Taken together, the training objective of our framework is formulated as,
	\begin{equation} \label{eq:6}
	\min\limits_{G, \mathcal{M}} \max\limits_{D, D_1, D_2, D_3} \mathcal{L}_{G} + \alpha ({\mathcal{L}_{adv}^{s}} + {\mathcal{L}_{adv}^{t{\rightarrow}s}})  + \beta (\mathcal{L}_{rec}^{s} + \mathcal{L}_{rec}^{t{\rightarrow}s})
	\end{equation} 
	where $ \alpha $ and $ \beta $ leverage the importance of losses above. Notably, our method is able to implicitly encourage $ G $ to generate semantic-consistent segmentation labels for the target domain.

	\section{Experiments}
	In this section, a comprehensive evaluation is performed on two domain adaption tasks to assess our framework for semantic segmentation. Specifically, we consider the large distribution shift of adapting from synthetic (i.e., GTA5 \cite{richter2016playing} and SYNTHIA \cite{ros2016synthia}) to the real images in Cityscapes \cite{cordts2016cityscapes}. A thorough comparison with the state-of-the-art methods and extensive ablation studies are also carried out to verify the effectiveness of each component in our framework. 
	
	\begin{table*}[t]
		\caption{A performance comparison of our method with other state-of-the-art models on "GTA5 to Cityscapes". The performance is measured by the intersection-over-union (IoU) for each class and mean IoU (mIoU). Two base architectures, i.e., VGG16 (V) and ResNet101 (R) are used in our study.}
		\label{table:gta2city}
		
		\tiny
		\setlength\tabcolsep{0.5pt}
		\begin{center}
			\begin{tabular}{ @{} l|c|*{19}{c}|*{1}{c} @{} }
				\toprule
				\multicolumn{22}{ c }{\bf GTA5$\,\to\,$Cityscapes } \\
				\midrule
				& \rotatebox[origin=c]{90}{Base} & \rotatebox[origin=c]{90}{road} & \rotatebox[origin=c]{90}{sidewalk} & \rotatebox[origin=c]{90}{building} & \rotatebox[origin=c]{90}{wall} & \rotatebox[origin=c]{90}{fence} & \rotatebox[origin=c]{90}{pole} & \rotatebox[origin=c]{90}{  traffic light  } & \rotatebox[origin=c]{90}{traffic sign} & \rotatebox[origin=c]{90}{vegetation} & \rotatebox[origin=c]{90}{terrain} & \rotatebox[origin=c]{90}{sky} & \rotatebox[origin=c]{90}{person} & \rotatebox[origin=c]{90}{rider} & \rotatebox[origin=c]{90}{car} & \rotatebox[origin=c]{90}{truck} & \rotatebox[origin=c]{90}{bus} & \rotatebox[origin=c]{90}{train} & \rotatebox[origin=c]{90}{motorbike} & \rotatebox[origin=c]{90}{bicycle} & \rotatebox[origin=c]{90}{\bf mIoU } \\ 
				\midrule
				
				Source only & R &
				75.8 & 16.8 & 77.2 & 12.5 & 21.0 & 25.5 & 30.1 & 20.1 & 81.3 & 24.6 & 
				70.3 & 53.8 & 26.4 & 49.9 & 17.2 & 25.9 & 6.5 & 25.3 & 36.0 & 36.6 \\
				
				
				SIBAN \cite{luo2019significance} & R &
				88.5 & 35.4 & 79.5 & 26.3 & 24.3 & 28.5 & 32.5 & 18.3 & 81.2 & 40.0 & 76.5 & 
				58.1 & 25.8 & 82.6 & 30.3 & 34.4 & 3.4 & 21.6 & 21.5 & 42.6 \\ 
				
				CLAN \cite{luo2019taking} & R &
				87.0 & 27.1 & 79.6 & 27.3 & 23.3 & 28.3 & 35.5 & 24.2 & 83.6 & 27.4 & 74.2 & 
				58.6 & 28.0 & 76.2 & 33.1 & 36.7 & 6.7 & 31.9 & 31.4 & 43.2 \\
				
				DISE \cite{chang2019all} & R &
				91.5 & 47.5 & 82.5 & 31.3 & 25.6 & 33.0 & 33.7 & 25.8 & 82.7 & 28.8 & 82.7 & 
				62.4 & 30.8 & 85.2 & 27.7 & 34.5 & 6.4 & 25.2 & 24.4 & 45.4 \\
				
			    IntraDA \cite{pan2020unsupervised} & R &
				90.6 & 37.1 & 82.6 & 30.1 & 19.1 & 29.5 & 32.4 & 
				20.6 & \bf 85.7 & 40.5 & 79.7 & 58.7 & 31.1 & \bf 86.3 & 
				31.5 & 48.3 & 0.0 & 30.2 & 35.8 & 46.3  \\
				
				BDL \cite{li2019bidirectional} & R &
				91.0 & 44.7 & 84.2 & 34.6 & 27.6 & 30.2 & 36.0 & 36.0 & 85.0 & \bf43.6 &
				83.0 & 58.6 & 31.6 & 83.3 & 35.3 & 49.7 & 3.3 & 28.8 & 35.6 & 48.5 \\
				
				CrCDA \cite{huang2020contextual}  & R &
				92.4 & \bf 55.3 & 82.3 & 31.2 & 29.1 & 32.5 & 33.2 &  35.6 & 83.5 & 34.8 & 84.2 & 58.9 & 32.2 & 84.7 &  40.6 & 46.1 & 2.1 & 31.1 & 32.7 & 48.6 \\
				
				SIM \cite{wang2020differential} & R &
				90.6 & 44.7 & 84.8 & 34.3 & 28.7 & 31.6 & 35.0 & 
				37.6 & 84.7 & 43.3 & 85.3 & 57.0 & 31.5 & 83.8 & 
				42.6 & 48.5 & 1.9 & 30.4 & 39.0 & 49.2 \\
				
				Kim et al. \cite{kim2020learning} & R &
				\bf 92.9 & 55.0 & \bf 85.3 & 34.2 & \bf 31.1 & 34.9 & \bf 40.7 & 
				34.0 & 85.2 & 40.1 & \bf 87.1 & 61.0 & 31.1 & 82.5 & 
				32.3 & 42.9 & 0.3 & \bf 36.4 & 46.1 & 50.2 \\
				
				FDA-MBT \cite{yang2020fda} & R &
				92.5 & 53.3 & 82.4 & 26.5 & 27.6 & \bf 36.4 & 40.6 & 
				\bf 38.9 & 82.3 & 39.8 & 78.0 & \bf 62.6 & \bf 34.4 & 84.9 & 
				34.1 & \bf 53.1 & \bf 16.9 & 27.7 & \bf 46.4 & \bf 50.45 \\
				
				\midrule
				
				Ours & R &
				90.8 & 41.4 & 84.7 & \bf 35.1 &27.5&31.2&38.0&32.8&85.6&42.1&84.9&59.6&
				\bf34.4&85.0&\bf42.8&52.7&3.4&30.9&38.1&49.5 \\
				
				\midrule                    
				\midrule
				
				Source only & V &
				26.0 & 14.9 & 65.1 & 5.5 & 12.9 & 8.9 & 6.0 & 2.5 & 70.0 & 2.9 & 47.0 & 24.5 & 
				0.0 & 40.0 & 12.1 & 1.5 & 0.0 & 0.0 & 0.0 & 17.9 \\
				
				SIBAN \cite{luo2019significance} & V &
				83.4 & 13.0 & 77.8 & 20.4 & 17.5 & 24.6 & 22.8 & 9.6 & 81.3 & 29.6 & 77.3 &
				42.7 & 10.9 & 76.0 & 22.8 & 17.9 & 5.7 & 14.2 & 2.0 & 34.2 \\
				
				ASN \cite{tsai2018learning} & V &
				87.3 & 29.8 & 78.6 & 21.1 & 18.2 & 22.5 & 21.5 & 11.0 & 79.7 & 29.6 &
				71.3 & 46.8 & 6.5 & 80.1 & 23.0 & 26.9 & 0.0 & 10.6 & 0.3 & 35.0 \\
				
				CyCADA \cite{hoffman2017cycada} & V &
				85.2 & 37.2 & 76.5 & 21.8 & 15.0 & 23.8 & 22.9 & 21.5 & 80.5 & 31.3 &
				60.7 & 50.5 & 9.0 & 76.9 & 17.1 & 28.2 & 4.5 & 9.8 & 0.0 & 35.4 \\    
				
				CLAN \cite{luo2019taking} & V &
				88.0 & 30.6 & 79.2 & 23.4 & 20.5 & 26.1 & 23.0 & 14.8 & 81.6 & 34.5 & 72.0 &
				45.8 & 7.9 & 80.5 & 26.6 & 29.9 & 0.0 & 10.7 & 0.0 & 36.6 \\
				
				CrDoCo \cite{chen2019crdoco} & V &
				89.1 & 33.2 & 80.1 & 26.9 & 25.0 & 18.3 & 23.4 & 12.8 & 77.0 & 29.1 & 72.4 & \bf55.1 & 20.2 & 79.9 & 22.3 & 19.5 & 1.0 & 20.1 & 18.7 & 38.1 \\
				
				CrCDA \cite{huang2020contextual} & V &
				86.8 & 37.5 & 80.4 & 30.7 & 18.1 & 26.8 & 25.3 & 
				15.1 & 81.5 & 30.9 & 72.1 & 52.8 & 19.0 & 82.1 & 
				25.4 & 29.2 & 10.1 & 15.8 & 3.7 & 39.1 \\
				
				BDL \cite{li2019bidirectional} & V &
				89.2 & 40.9 & 81.2 & 29.1 & 19.2 & 14.2 & 29.0 & 19.6 & 83.7 & 35.9 &
				80.7 & 54.7 & 23.3 & 82.7 & 25.8 & 28.0 & 2.3 & 25.7 & 19.9 & 41.3 \\  
				
				FDA-MBT \cite{yang2020fda} & V &
				86.1 & 35.1 & 80.6 & 30.8 & 20.4 & 27.5 & 30.0 & 
				26.0 & 82.1 & 30.3 & 73.6 & 52.5 & 21.7 & 81.7 & 
				24.0 & 30.5 & \bf 29.9 & 14.6 & 24.0 & 42.2 \\
				
				Kim et al. \cite{kim2020learning} & V &
				\bf 92.5 & \bf 54.5 & \bf 83.9 & \bf 34.5 & \bf 25.5 & \bf 31.0 & 30.4 & 
				18.0 & \bf 84.1 & \bf 39.6 & \bf 83.9 & 53.6 & 19.3 & 81.7 & 
				21.1 & 13.6 & 17.7 & 12.3 & 6.5 & 42.3 \\
				
				SIM \cite{wang2020differential} & V &
				88.1 & 35.8 & 83.1 & 25.8 & 23.9 & 29.2 & 28.8 & 
				\bf 28.6 & 83.0 & 36.7 & 82.3 & 53.7 & 22.8 & 82.3 & 
				26.4 & \bf 38.6 & 0.0 & 19.6 & 17.1 & 42.4 \\
				
				\midrule
				
				Ours & V &
				90.1&41.2&82.2&30.3&21.3&18.3&\bf33.5&23.0&\bf84.1&37.5&81.4&54.2&\bf24.3&\bf83.0&\bf27.6&32.0&8.1&\bf29.7&\bf26.9&\bf43.6 \\    
				
				\bottomrule
			\end{tabular}
		\end{center}
	\end{table*}

	\subsection{Datasets}
	Cityscapes is one of the benchmarks for urban scene understanding, which is collected from 50 cities with varying scene layouts and weather conditions. The 5,000 finely-annotated images from this dataset are used in our study, which contains 2,975 training images, 500 validation images, and 1,525 test images. Each image with a resolution of 2048 $ \times $ 1024. The GTA5 dataset is synthesized from the game Grand Theft Auto V (GTA5), including a total of 24,966 labeled images whose annotations are compatible with Cityscapes. The resolution of each image is 1914 $ \times $ 1052. The SYNTHIA-RAND-CITYSCAPES (or SYNTHIA for short) contains 9,400 pixel-level annotated images (1280 $ \times $ 760), which are synthesized from a virtual city. Following the same setting reported in the previous studies, we use the labeled SYNTHIA or GTA5 dataset as the source domain, while using the unlabeled training dataset in the CITYSCAPES as the target domain. Only the 500 labeled validation images from CITYSCAPES are used as test data in all of our experiments.
	
	\subsection{Network Architecture}
	We use two segmentation baseline models, i.e., FCN-VGG16 and DeepLab-ResNet101 to investigate the effectiveness and generalizability of our framework. Specifically, FCN-VGG16 is the combination of FCN-8s \cite{long2015fully} and VGG16 \cite{simonyan2014very}, while DeepLab-ResNet101 is obtained by integrating DeepLab-V2 \cite{chen2018deeplab} into ResNet101 \cite{he2016deep}. These two segmentation models share the same discriminator which has 5 convolution layers with channel number {64,128, 256, 512, 1}. For each layer, a leaky ReLU parameterized by 0.2 is followed, except the last one. The kernel size and stride are set to 4$ \times $4 and 2, respectively. The reconstruction model follows the architecture in \cite{johnson2016perceptual}, containing 3 convolution layers (kernel 3$ \times $3 and stride 1), 9 ResNet blocks (kernel 3$ \times $3 and stride 2), and another 3 transposed convolution layers (kernel 3$ \times $3 and stride 2) for upsampling. The 3 multi-scale discriminators share the identical network, each of which follows the architecture of PatchGAN \cite{isola2017image}. More details regarding the architecture of discriminators in both segmentation and reconstruction models can be found in the \textbf{Supplementary}.
	
	
	\begin{table*}[t]
		\caption{A performance comparison of our method with other state-of-the-art models on "SYNTHIA to Cityscapes". The performance is measured by the IoU for each class and mIoU. Two base architectures, i.e., VGG16 (V) and ResNet101 (R) are used in our study.}
		\label{table:synthia2city}
		
		\tiny
		\setlength\tabcolsep{1.8pt}
		\begin{center}
			\begin{tabular}{ @{} l|c|*{16}{c}|*{1}{c} @{} }
				\toprule
				\multicolumn{19}{ c }{\bf SYNTHIA$\,\to\,$Cityscapes } \\
				\midrule
				& \rotatebox[origin=c]{90}{Base} & \rotatebox[origin=c]{90}{road} & \rotatebox[origin=c]{90}{sidewalk} & \rotatebox[origin=c]{90}{building} & \rotatebox[origin=c]{90}{wall} & \rotatebox[origin=c]{90}{fence} &\rotatebox[origin=c]{90}{pole} & \rotatebox[origin=c]{90}{traffic light} & \rotatebox[origin=c]{90}{traffic sign} & \rotatebox[origin=c]{90}{vegetation} & \rotatebox[origin=c]{90}{sky} & \rotatebox[origin=c]{90}{person} & \rotatebox[origin=c]{90}{rider} & \rotatebox[origin=c]{90}{car} & \rotatebox[origin=c]{90}{bus} & \rotatebox[origin=c]{90}{motorbike} & \rotatebox[origin=c]{90}{bicycle} & \rotatebox[origin=c]{90}{\bf mIoU} \\ 
				\midrule
				
				Source only & R &
				55.6 & 23.8 & 74.6 & \textemdash & \textemdash & \textemdash & 6.1 & 12.1 & 74.8 & 79.0 & 55.3 & 19.1 & 39.6 & 23.3 & 13.7 & 25.0 & 38.6 \\                    
				
				ASN \cite{tsai2018learning} & R &
				84.3 & 42.7 & 77.5 & \textemdash & \textemdash & \textemdash & 4.7 & 7.0 & 77.9 & 82.5 & 54.3 & 21.0 & 72.3 & 32.2 & 
				18.9 & 32.3 & 46.7 \\
				
				
				DISE \cite{chang2019all} & R &
				91.7 & \bf 53.5 & 77.1 & \textemdash & \textemdash & \textemdash & 6.2 & 7.6 & 78.4 & 81.2 & 55.8 & 19.2 & 82.3 & 30.3 & 
				17.1 & 34.3 & 48.8\\
				
				IntraDA \cite{pan2020unsupervised} & R &
				84.3 & 37.7 & 79.5 & \textemdash & \textemdash & \textemdash & 9.2 & 8.4 & 80.0 & 84.1 & 57.2 & 
				23.0 & 78.0 & 38.1 & 20.3 & 36.5 & 48.9 \\
				
				Kim et al. \cite{kim2020learning} & R &
				\bf 92.6 & 53.2 & 79.2 & \textemdash & \textemdash & \textemdash & 1.6 & 7.5 & 78.6 & 84.4 & 52.6 & 
				20.0 & 82.1 & 34.8 & 14.6 & 39.4 & 49.3 \\
				
				DADA \cite{vu2019dada} & R &
				89.2 & 44.8 & \bf 81.4 & \textemdash & \textemdash & \textemdash & 8.6 & 11.1 & \bf81.8 & 84.0 & 54.7 & 19.3 & 79.7 & 40.7 & 14.0 & 38.8 & 49.8 \\
				
				CrCDA \cite{huang2020contextual} & R &
				86.2 & 44.9 & 79.5 & \textemdash & \textemdash & \textemdash & 9.4 & 11.8 & 78.6 & \bf 86.5 & 57.2 & 
				26.1 & 76.8 & 39.9 & 21.5 & 32.1 & 50.0 \\
				
				BDL \cite{li2019bidirectional} & R &
				86.0 & 46.7 & 80.3 & \textemdash & \textemdash & \textemdash & 14.1 & 11.6 & 79.2 & 81.3 & 54.1 & 27.9 & 73.7 & \bf42.2 & 25.7 & 45.3 & 51.4 \\
				
				SIM \cite{wang2020differential} & R &
				83.0 & 44.0 & 80.3 & \textemdash & \textemdash &  \textemdash & 17.1 & 15.8 & 80.5 & 81.8 & 
				59.9 & \bf 33.1 & 70.2 & 37.3 & 28.5 & 45.8 & 52.1 \\
				
				FDA-MBT \cite{yang2020fda} & R &
				79.3 & 35.0 & 73.2 & \textemdash & \textemdash &  \textemdash & \bf 19.9 & \bf 24.0 & 61.7 & 82.6 & \bf 61.4 & 
				31.1 & \bf 83.9 & 40.8 & \bf 38.4 & \bf 51.1 & 52.5 \\
				\midrule
				
				Ours & R &
				85.1&44.5&81.0& \textemdash & \textemdash & \textemdash&16.4&15.2&80.1&84.8&59.4&31.9&73.2&41.0&32.6&44.7&\bf53.1 \\
				
				\midrule
				\midrule
				CrCDA \cite{huang2020contextual} & V &
				74.5 & 30.5 & 78.6 & 6.6 & \bf 0.7 & 21.2 & 2.3 & 
				8.4 & 77.4 & 79.1 & 45.9 & 16.5 & 73.1 & 24.1 & 
				9.6 & 14.2 & 35.2 \\
				
				ROAD-Net \cite{chen2018road} & V &
				77.7 & 30.0 & 77.5 & 9.6 & 0.3 & 25.8 & 10.3 & 15.6 & 77.6 & 79.8 & 44.5 & 
				16.6 & 67.8 & 14.5 & 7.0 & 23.8 & 36.2 \\
				
				SPIGAN \cite{lee2018spigan} & V &
				71.1 & 29.8 & 71.4 & 3.7 & 0.3 & \bf33.2 & 6.4 & 15.6 & 81.2 & 78.9 & 52.7 & 
				13.1 & 75.9 & 25.5 & 10.0 & 20.5 & 36.8 \\
				
				GIO-Ada \cite{chen2019learning} & V &
				78.3 & 29.2 & 76.9 & \bf11.4 & 0.3 & 26.5 & 10.8 & 17.2 & 81.7 & \bf81.9 & 45.8 & 
				15.4 & 68.0 & 15.9 & 7.5 & 30.4 & 37.3 \\
				
				TGCF-DA \cite{Choi2019self} & V &
				\bf90.1 & \bf48.6 & \bf80.7 & 2.2 & 0.2 & 27.2 & 3.2 & 14.3 & \bf82.1 & 78.4 & 54.4 & 16.4 & \bf82.5 & 12.3 & 1.7 & 21.8 & 38.5 \\
				
				BDL \cite{li2019bidirectional} & V &
				72.0 & 30.3 & 74.5 & 0.1 & 0.3 & 24.6 & 10.2 & 25.2 & 80.5 & 80.0 &
				54.7 & 23.2 & 72.7 & 24.0 & 7.5 & 44.9 & 39.0 \\
				
				FDA-MBT \cite{yang2020fda} & V &
				84.2 & 35.1 & 78.0 & 6.1 & 0.44 & 27.0 & 8.5 & 
				22.1 & 77.2 & 79.6 & 55.5 & 19.9 & 74.8 & 24.9 & 
				\bf 14.3 & 40.7 & 40.5 \\
				
				\midrule
				Ours & V &
				73.7&29.6&77.6&1.0&0.4&26.0&\bf14.7&\bf26.6&80.6&81.8&\bf57.2&\bf24.5&76.1&\bf27.6&13.6&\bf46.6&
				\bf41.1 \\
				
				\bottomrule
			\end{tabular}
		\end{center}
	\end{table*}
	
	\subsection{Implementation Details}
	Our framework is implemented with PyTorch \cite{paszke2017automatic} on two TITAN Xp GPUs, each of which with 12GB memory. The batch size is set to one for training all the models discussed above. Limited by the GPU memory space, the translation network is first trained to perform target-to-source image translation by using Adam optimizer \cite{kingma2014adam}. The initial learning rate is set to 0.0001, which is reduced by half after every 100,000 iterations. We use momentum \{0.5, 0.999\} with weight decay 0.0001. The maximum training iteration is 1000$ k $. 
	
	\newcommand{\cmark}{\ding{51}}%
	\begin{table}[t]
		\caption{Ablation study of the target-to-source translation and the reconstruction network. S$\,\to\,$T and T$\,\to\,$S indicate source-to-target and target-to-source translation. }
		\label{table:ablation_gta2city}
		
		\footnotesize
		\setlength\tabcolsep{5pt}
		\begin{center}
			\begin{tabularx}{.73\textwidth}{ cccc|c|c @{} }
				\toprule
				Base & S$\,\to\,$T &  T$\,\to\,$S & Reconstruction & GTA5 & SYNTHIA \\
				\midrule
				R & \cmark & & & 48.5 & 51.4 \\
				R & & \cmark & & 49.1 & 52.0 \\
				R & & \cmark & \cmark & 49.5 & 53.1 \\                       
				\midrule
				
				V & \cmark & & & 41.3 & 39.0 \\
				V & & \cmark & & 42.3 & 40.1 \\
				V & & \cmark & \cmark & 43.6 & 41.1 \\
				
				\bottomrule
			\end{tabularx}
		\end{center}
	\end{table}
	
	\begin{figure*}
		\begin{center}
			\includegraphics[width=1.0\linewidth]{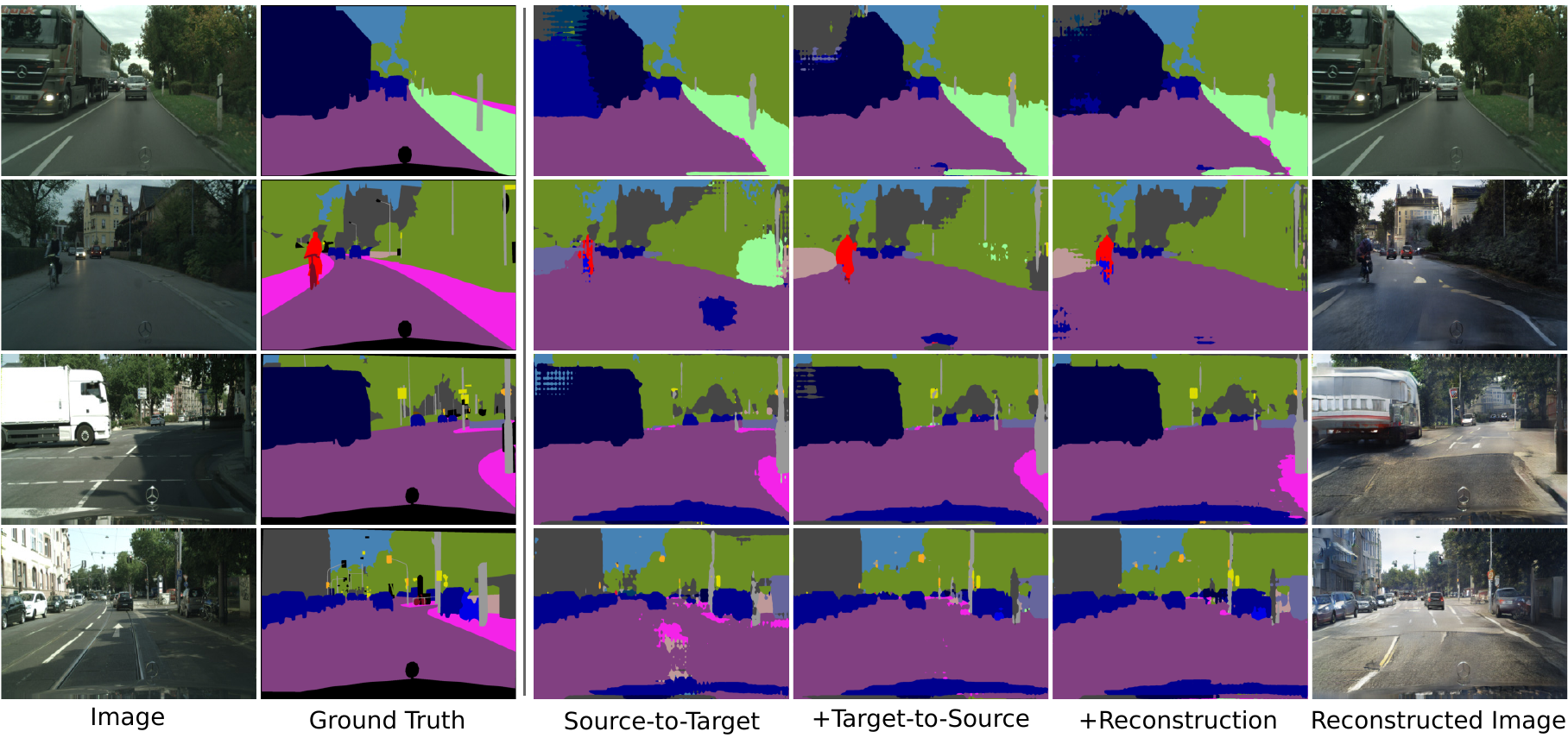}
		\end{center}
		\caption{Qualitative examples of semantic segmentation results in Cityscapes. For each target-domain image (first column), its ground truth and the corresponding segmentation prediction from the baseline model (source-to-target translation) are given. The following are predictions of our method by incorporating target-to-source translation and reconstruction, together with the reconstructed image.}
		\label{fig:quali_gta2city}
	\end{figure*}
	
	DeepLab-ResNet101 is trained using Stochastic Gradient Descent optimizer with initial learning rate $ 2.5\times10^{-4} $. The polynomial decay with power 0.9 is applied to the learning rate. The momentum and weight decay are set to 0.9 and $ 5\times10^{-4} $, respectively. For FCN-VGG16, the Adam optimizer with momentum \{0.9, 0.99\} and initial learning rate $ 1\times10^{-5} $ is used for training. The learning rate is decreased using step decay with step size 50000 and drop factor 0.1. In equation \ref{eq:1}, $ \lambda $ is set to $ 1\times10^{-3} $ for DeepLab-ResNet101 and $ 1\times10^{-4} $ for FCN-VGG16.
	
	The reconstruction network is first pre-trained by reconstructing source images $ \mathcal{X}_{s} $ from the corresponding labels $ Y_s $. We use the Adam optimizer with initial learning rate $ 2\times10^{-4} $ and momentum \{0.5, 0.999\}, where the learning rate is linearly decreased to zero. In equation \ref{eq:6}, we set $ \beta=10 $. $ \alpha $ is set to $ 0.01 $ and $ 0.001 $ for FCN-VGG16 and DeepLab-ResNet101 respectively.

	\subsection{GTA5$\,\to\,$Cityscapes}
	
	We carry out the adaptation from GTA5 to Cityscapes by following the same evaluation protocol as previously reported in \cite{tsai2018learning,li2019bidirectional}. The overall quantitative performance is assessed on 19 common classes (e.g., road, wall, and car) between these two domains. As shown in Table~\ref{table:gta2city}, we demonstrate competitive performance against ResNet101-based methods, but are inferior to two newly published models \cite{kim2020learning,yang2020fda}. For the VGG16-based backbone, however, we are able to achieve the best results compared to existing state-of-the-art methods including \cite{kim2020learning,yang2020fda}. Specifically, our method surpasses the source-only model (without adaptation) by 12.9\% and 25.7\% on ResNet101 and VGG16, respectively. Compared with CyCADA \cite{hoffman2017cycada} and BDL \cite{li2019bidirectional} that rely on source-to-target translation, we demonstrate significant improvements (i.e., 8.2\% and 2.3\% on VGG16) by reducing image translation bias. CLAN \cite{luo2019taking} aims to enforce local semantic consistency by a category-level adversarial network. However, such a strategy fails to account for the global semantic consistency. Our reconstruction network shares a similar spirit with CLAN in terms of joint distribution alignment but enables us to enforce semantic consistency from a global view. As a consequence, we get 6.3\% and 7.0\% improvement on ResNet101 and VGG16, respectively. 
	
	\subsection{SYNTHIA$\,\to\,$Cityscapes}
	
	We then evaluate our framework on the adaptation from SYNTHIA to Cityscapes based on 13 classes on ResNet101 and 16 classes on VGG16. The results exhibit that our method outperforms other competing methods on average as shown in Table~\ref{table:synthia2city}. Both ASN \cite{tsai2018learning} and BDL \cite{li2019bidirectional} adapt output space in their models. However, simply aligning segmentation outputs may lead to negative transfer issue, owing to the dramatic differences of the layout and structure between SYNTHIA and Cityscapes. We achieve 6.4\% and 1.7\% improvement than ASN and BDL on ResNet101, respectively. It is noteworthy that we also outperform \cite{yang2020fda} on both ResNet101 and VGG16-based backbone.

	\begin{table}[t]
		\caption{Ablation study of the temperature $ \tau $ on GTA5$\,\to\,$Cityscapes.}
		\label{table:ablation_temperature}
		
		\footnotesize
		\setlength\tabcolsep{10pt}
		\begin{center}
			\begin{tabularx}{.7\textwidth}{ cccccc @{} }
				\toprule
				$ \tau $ & 0.0001 & 0.001 & 0.01 & 0.1 & 1 \\
				\midrule
				mIoU & 42.7 & \bf 43.6 & 42.8 & 42.9 & 41.5 \\
				\bottomrule
			\end{tabularx}
		\end{center}
	\end{table}
	
	\subsection{Ablation Study}
	
	{\textbf{Target-to-source Translation and Reconstruction}} For GTA5 to Cityscapes, 0.6\% improvement is achieved by considering target-to-source translation on ResNet101 compared to the source-to-target translation model (Table~\ref{table:ablation_gta2city}). By further enforce semantic consistency through a reconstruction network, our method achieves 49.5 mIoU. Similar improvements are also observed on VGG16, with 1.0\% improvement by performing target-to-source translation. The prediction power of our method is further boosted by combining translation and reconstruction, giving rise to another 1.3\% mIoU improvement. The qualitative study of each module in our method is showcased in Figure~\ref{fig:quali_gta2city}.

	\begin{table}[t]
		\caption{Ablation study of the feature space vs. label space reconstruction.}
		\label{table:ablation_recons}
		
		\footnotesize
		\setlength\tabcolsep{10pt}
		\begin{center}
			\begin{tabularx}{.75\textwidth}{ ccc @{} }
				\toprule
				 & Feature space & Label space \\
				\midrule
				GTA5$\,\to\,$Cityscapes & 41.48 & 43.6 \\
				\midrule
				SYNTHIA$\,\to\,$Cityscapes & 40.13 & 41.1 \\
				\bottomrule
			\end{tabularx}
		\end{center}
		\vspace{-0.2in}
	\end{table}

	\begin{wraptable}{r}{0.5\textwidth}
		\caption{Ablation study of the reconstruction loss on GTA5$\,\to\,$Cityscapes with VGG16 backbone.}
		\label{table:ablation_recons_loss}
		
		\footnotesize
		\setlength\tabcolsep{6pt}
		\begin{center}
			\begin{tabularx}{.4\textwidth}{ cc|c @{} }
				\toprule
				VGG & Discriminator & mIoU \\
				\midrule
				& & 41.53 \\
				\cmark & & 42.82 \\
				& \cmark & 41.95 \\
				\cmark & \cmark & 43.6 \\
				\bottomrule
			\end{tabularx}
		\end{center}
	\end{wraptable}
	
	For SYNTHIA to Cityscapes, we achieve a performance boost of 0.6\% and 1.1\% by considering target-to-source translation on ResNet101 and VGG16, respectively (Table~\ref{table:ablation_gta2city}). The performance gain is 1.1\% and 1.0\% by incorporating the reconstruction network. Our results prove the effectiveness of target-to-source translation and reconstruction in adapting domain knowledge for semantic segmentation.
	
	{\textbf{Parameter Analysis}} We investigate the sensitivity of temperature parameter $ \tau $ in this section and find that $ \tau=0.001 $ achieves the best performance (Table~\ref{table:ablation_temperature}). Therefore, $ \tau $ is set to $ 0.001 $ in all of our experiments to approximate semantic label maps. 
	
	{\textbf{Feature Space VS. Label Space Reconstruction}} We also evaluate the feature space reconstruction based on the VGG16-based backbone. Table~\ref{table:ablation_recons} highlights the benefits of our label-driven reconstruction that enforces semantic consistency of target images and their predicted labels.
	
	{\textbf{Reconstruction loss}} Table~\ref{table:ablation_recons_loss} shows the complementary role of VGG perceptual loss and discriminator feature matching loss (equation \ref{eq:recons}) in maintaining semantic consistency.

	\section{Conclusion}
	We propose a novel framework that exploits cross-domain adaptation in the context of semantic segmentation. Specifically, we translate images from the target domain to the source domain to reduce image translation bias and the computational cost. To enforce cross-domain features with the same category close to each other, we reconstruct both source and target images directly from the label space. Experiments demonstrate that our method achieves significant improvement in adapting from GTA5 and SYNTHIA to Cityscapes.
	
	\paragraph{\textbf{Acknowledgments}}
	This work was partially supported by US National Science Foundation IIS-1718853, the CAREER grant IIS-1553687 and Cancer Prevention and Research Institute of Texas (CPRIT) award (RP190107).
\clearpage
%
%
\bibliographystyle{splncs04}
\bibliography{5818}
\end{document}



\section{Supplementary}
\subsection{Target-to-source Translation}

Inspired by image-to-image translation networks \cite{zhu2017unpaired,liu2017unsupervised}, existing domain adaptation methods \cite{hoffman2017cycada,murez2017image,li2019bidirectional} translate images from the source domain to the target domain (source-to-target) to reduce pixel-level domain discrepancy. This is achieved by an unsupervised image translation model $ \mathcal{F}^{-1} $ such as CycleGAN \cite{zhu2017unpaired} to learn a mapping $ \mathcal{F}^{-1}: \mathcal{X}_s \rightarrow \mathcal{X}_t $. However, such strategy introduces inevitable bias to the translated images $ \mathcal{F}^{-1}(\mathcal{X}_s) $, stemming from that $ \mathcal{F}^{-1}(\mathcal{X}_s) $ and $ \mathcal{X}_t $ cannot be guaranteed to follow the exactly identical distribution through the adversarial learning \cite{goodfellow2014generative}. This problem can get even worse in the source-to-target translation, as $ |\mathcal{X}_s| \gg |\mathcal{X}_t| $ in most of domain adaptation problems. For example, GTA5 \cite{richter2016playing} (i.e., $ \mathcal{X}_s $) contains 24,966 images, while Cityscapes (i.e., $ \mathcal{X}_t $) \cite{cordts2016cityscapes} has only 2,975 images. As a consequence, $ \mathcal{F}^{-1}(\mathcal{X}_s) $ contains massive amounts of translation bias which will further induces negative effects when adapting domain knowledge between $ \mathcal{F}^{-1}(\mathcal{X}_s) $ and $ \mathcal{X}_t $. To alleviate this problem, for the first time, we translate images from the target domain to the source domain through the mapping $ \mathcal{F}: \mathcal{X}_t \rightarrow \mathcal{X}_s $ (where $ \mathcal{F} $ is the reverse function of $ \mathcal{F}^{-1} $). $ \mathcal{X}_s $ and $ \mathcal{F}(\mathcal{X}_t) $ are then used for further domain knowledge transfer. Doing so significantly reduces the translation bias in the translated images $ \mathcal{F}(\mathcal{X}_t) $ and is much more computationally efficient than the source-to-target translation. Another advantage is that the segmentation network can be trained using original source images $ \mathcal{X}_s $ and their corresponding labels.

\begin{table*}[t]
	\caption{A performance comparison of our method with other state-of-the-art models on "GTA5 to Cityscapes". The performance is measured by the mIoU gap between each model and the fully-supervised model (Oracle). Two base architectures, i.e., VGG16 (V) and ResNet101 (R) are used in our study.}
	\label{table:gta2city}
	
	\tiny
	\setlength\tabcolsep{5pt}
	\begin{center}
		\begin{tabular}{ @{} l|c|c|c @{} }
			\toprule
			\multicolumn{4}{ c }{\bf GTA5$\,\to\,$Cityscapes } \\
			\midrule
			& {Base} & {\bf mIoU Gap} & {\bf Oracle } \\ 
			\midrule
			
			Source only & R &
			-28.5 & 65.1 \\
			
			
			SIBAN \cite{luo2019significance} & R &
			-22.5 & 65.1 \\ 
			
			CLAN \cite{luo2019taking} & R &
			-21.9 & 65.1 \\
			
			DISE \cite{chang2019all} & R &
			-19.7 & 65.1 \\
			
			IntraDA \cite{pan2020unsupervised} & R &
			-18.8 & 65.1 \\
			
			BDL \cite{li2019bidirectional} & R &
			-16.6 & 65.1 \\
			
			CrCDA \cite{huang2020contextual}  & R &
			-16.5 & 65.1 \\
			
			SIM \cite{wang2020differential} & R &
			-15.9 & 65.1 \\
			
			Kim et al. \cite{kim2020learning} & R &
			-14.9 & 65.1 \\
			
			FDA-MBT \cite{yang2020fda} & R &
			\bf -14.65 & 65.1 \\
			
			\midrule
			
			Ours & R &
			-15.6 & 65.1 \\
			
			\midrule                    
			\midrule
			
			Source only & V &
			-46.7 & 64.6 \\
			
			SIBAN \cite{luo2019significance} & V &
			-26.1 & 60.3 \\
			
			ASN \cite{tsai2018learning} & V &
			-25.2 & 61.8 \\
			
			CyCADA \cite{hoffman2017cycada} & V &			
			-24.9 & 60.3 \\    
			
			CLAN \cite{luo2019taking} & V &			
			-23.7 & 60.3 \\
			
			CrDoCo \cite{chen2019crdoco} & V &
			-22.2 & 60.3 \\
			
			CrCDA \cite{huang2020contextual} & V &
			-22.7 &  61.8 \\
			
			BDL \cite{li2019bidirectional} & V &
			-19.0 & 60.3 \\  
			
			FDA-MBT \cite{yang2020fda} & V &
			-18.1 & 60.3 \\
			
			Kim et al. \cite{kim2020learning} & V &
			-18.0 & 60.3 \\
			
			SIM \cite{wang2020differential} & V &
			-17.9 & 60.3 \\
			
			\midrule
			
			Ours & V &
			\bf-16.8 & 60.3 \\    
			
			\bottomrule
		\end{tabular}
	\end{center}
\end{table*}

\begin{table*}
	\caption{A performance comparison of our method with other state-of-the-art models on "SYNTHIA to Cityscapes". The performance is measured by the mIoU gap between each model and the fully-supervised model (Oracle). Two base architectures, i.e., VGG16 (V) and ResNet101 (R) are used in our study.}
	\label{table:synthia2city}
	
	\tiny
	\setlength\tabcolsep{5pt}
	\begin{center}
		\begin{tabular}{ @{} l|c|c|c @{} }
			\toprule
			\multicolumn{4}{ c }{\bf SYNTHIA$\,\to\,$Cityscapes } \\
			\midrule
			& {Base} &
			{\bf mIoU Gap} & {\bf Oracle} \\ 
			\midrule
			
			Source only & R &
			\textemdash & \textemdash \\                    
			
			ASN \cite{tsai2018learning} & R &
			-25.0 & 71.7 \\
			
			
			DISE \cite{chang2019all} & R &
			-22.9 & 71.7 \\
			
			IntraDA \cite{pan2020unsupervised} & R &
			-22.8 & 71.7 \\
			
			Kim et al. \cite{kim2020learning} & R &
			-22.4 & 71.7 \\
			
			DADA \cite{vu2019dada} & R &
			-21.9 & 71.7 \\
			
			CrCDA \cite{huang2020contextual} & R &
			-21.7 & 71.7 \\
			
			BDL \cite{li2019bidirectional} & R &
			-20.3 & 71.7 \\
			
			SIM \cite{wang2020differential} & R &
			-19.6 & 71.7 \\
			
			FDA-MBT \cite{yang2020fda} & R &
			-19.2 & 71.7 \\
			\midrule
			
			Ours & R &
			\bf-18.6 & 71.7 \\
			
			\midrule
			\midrule
			CrCDA \cite{huang2020contextual} & V &
			-28.9 & 64.1 \\
			
			ROAD-Net \cite{chen2018road} & V &
			-27.6 & 64.1 \\
			
			SPIGAN \cite{lee2018spigan} & V &
			-22.7 & 59.5 \\
			
			GIO-Ada \cite{chen2019learning} & V &
			-26.8 & 64.1 \\
			
			TGCF-DA \cite{Choi2019self} & V &
			-25.6 & 64.1 \\
			
			BDL \cite{li2019bidirectional} & V &
			-20.5 & 59.5 \\
			
			FDA-MBT \cite{yang2020fda} & V &
			-19.0 & 59.5 \\
			
			\midrule
			Ours & V &
			\bf-18.4 & 59.5 \\
			
			\bottomrule
		\end{tabular}
	\end{center}
\end{table*}

\subsection{Network Architecture}
The three multi-scale discriminators (i.e., $ D_1 $, $ D_2 $, and $ D_3 $) used in our reconstruction model follow the identical network architecture. Each of them is a $ 70\times70 $ PatchGAN \cite{isola2017image} containing five convolution layers with kernel number \{64, 128, 256, 512, 1\}. The kernel size for each layer is $ 4 \times 4 $. The first three layers use stride 2, while the last two layers with stride 1. A leaky ReLU parameterized by 0.2 is applied to the first four layers. We also apply BatchNorm to each layer, except the first and last one.

\subsection{Implementation Details}

For multi-scale discriminators, Adam optimizer with initial learning rate $ 2\times10^{-4} $ and momentum \{ 0.5, 0.999\} are used in our study. The learning rate is linearly decayed to zero with step size 100.

\subsection{mIoU Gap}

We investigate our model's ability in narrowing the mIoU gap between the model (Oracle) that is trained in a fully-supervised matter. 
Compared to existing state-of-the-art methods, we significantly recover the performance loss based on the ResNet101 backbone on GTA5 to Cityscapes. Although we are inferior to \cite{kim2020learning,yang2020fda} which are published simultaneously with our work, we outperforms these two methods on VGG16-based backbone by a large margin (Table~\ref{table:gta2city}).  
Similar improvements can also be observed on the adaptation from SYNTHIA to Cityscapes as shown in Table~\ref{table:synthia2city}.

\subsection{Qualitative Comparison}

\begin{figure*}
	\setlength{\abovecaptionskip}{0pt}
	\begin{center}
		\includegraphics[width=1.0\linewidth]{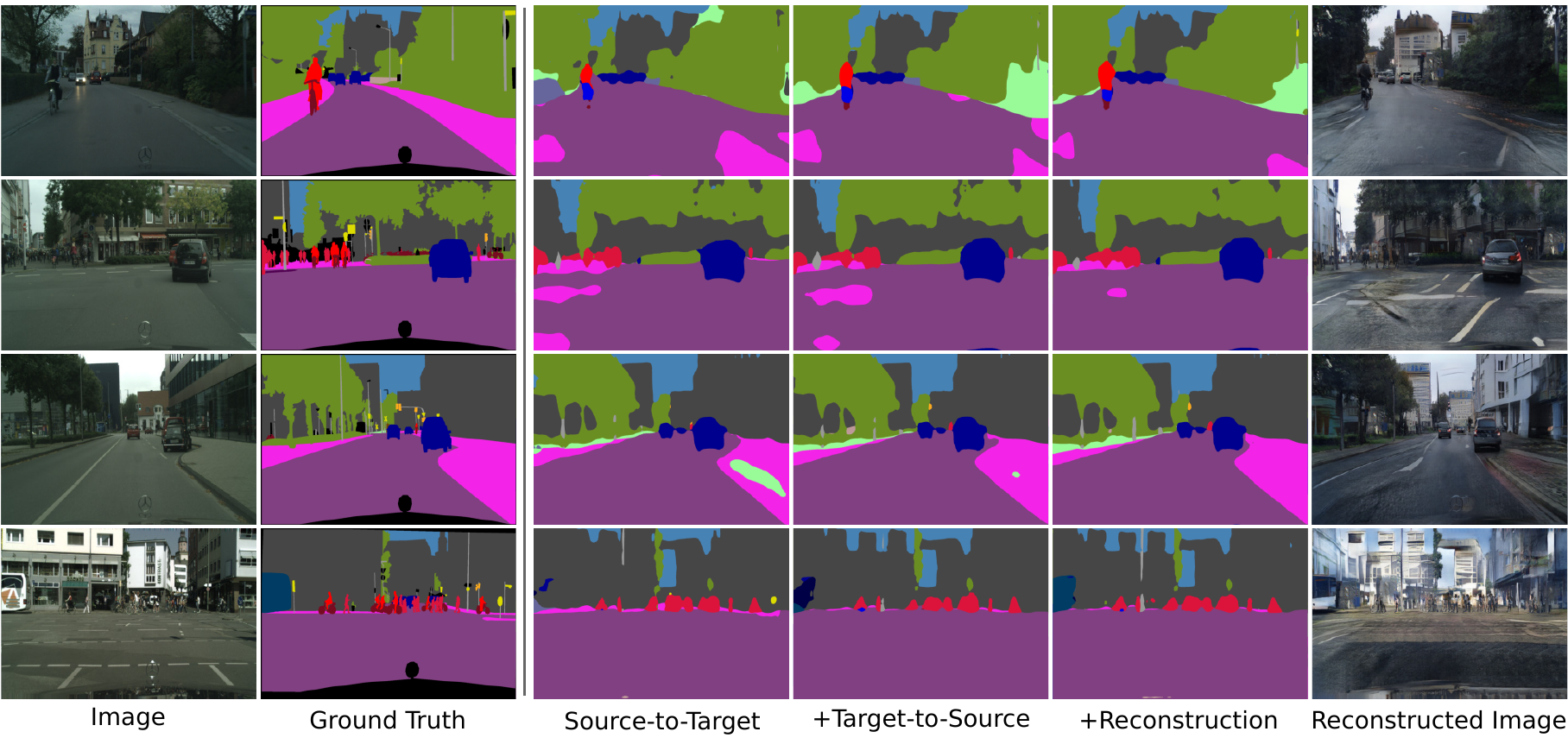}
	\end{center}
	\caption{Qualitative examples of semantic segmentation results in Cityscapes (GTA5$\rightarrow$Cityscapes with VGG16). For each target-domain image (first column), its ground truth and the corresponding segmentation output from the baseline model (source-to-target) are given. The following are predictions of our method by incorporating target-to-source translation and reconstruction, together with the reconstructed image.}
	\label{fig:quali_gta2city_vgg}
	\vspace{-0.1in}
\end{figure*}

\paragraph{\textbf{GTA5$\,\to\,$Cityscapes}}
As shown in Figure~\ref{fig:quali_gta2city_vgg}, we present the qualitative comparison in Cityscapes based on the VGG16 model from GTA5$\rightarrow$Cityscapes. Our results reveal the effectiveness of target-to-source translation and joint distribution alignment in adapting cross-domain knowledge.

\begin{figure*}
	\setlength{\abovecaptionskip}{0pt}
	\begin{center}
		\includegraphics[width=1.0\linewidth]{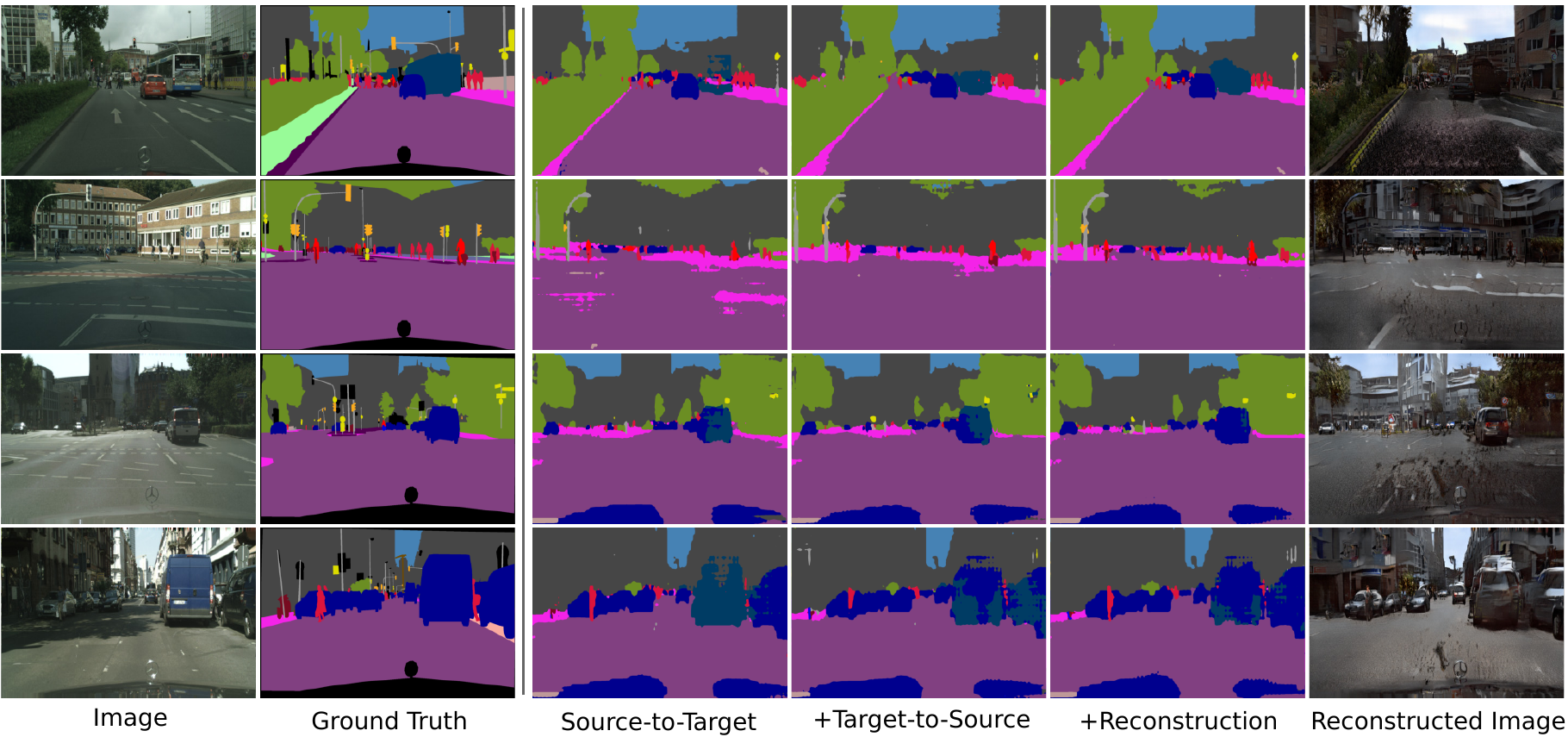}
	\end{center}
	\caption{Qualitative examples of semantic segmentation results in Cityscapes (SYNTHIA$\rightarrow$Cityscapes with ResNet101). For each target-domain image (first column), its ground truth and the corresponding segmentation output from the baseline model (source-to-target) are given. The following are predictions of our method by incorporating target-to-source translation and reconstruction, together with the reconstructed image.}
	\label{fig:quali_synthia2city_deeplab}
	\vspace{-0.2in}
\end{figure*}

\begin{figure*}
	\setlength{\abovecaptionskip}{0pt}
	\begin{center}
		\includegraphics[width=1.0\linewidth]{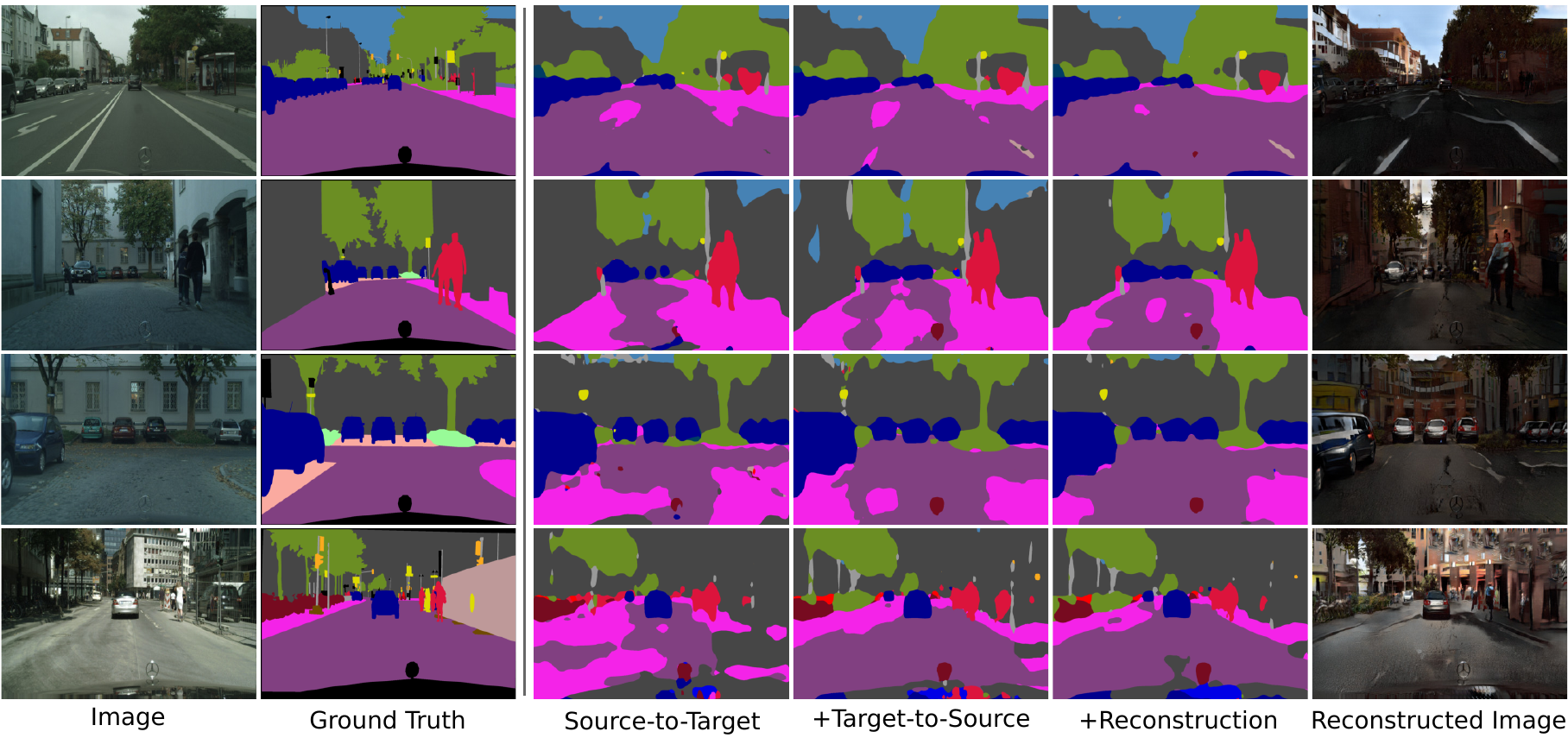}
	\end{center}
	\caption{Qualitative examples of semantic segmentation results in Cityscapes (SYNTHIA$\rightarrow$Cityscapes with VGG16). For each target-domain image (first column), its ground truth and the corresponding segmentation output from the baseline model (source-to-target) are given. The following are predictions of our method by incorporating target-to-source translation and reconstruction, together with the reconstructed image.}
	\label{fig:quali_synthia2city_vgg}
\end{figure*}

\paragraph{\textbf{SYNTHIA$\,\to\,$Cityscapes}}
The qualitative comparison for ResNet101 and VGG16 model from SYNTHIA$\rightarrow$Cityscapes are showcased in Figure~\ref{fig:quali_synthia2city_deeplab} and Figure~\ref{fig:quali_synthia2city_vgg}, respectively. Similarly, each component in our framework contributes to the overall performance improvement.

\vfill

{\small
\bibliographystyle{ieee_fullname}
\bibliography{egbib}
}